\documentclass[sigconf]{acmart}
\acmConference[Conference]{Z}{2026}{USA}
\usepackage{threeparttable} 
\usepackage{array}
\usepackage{subfigure}
\usepackage{tablefootnote}
\usepackage{booktabs}
\usepackage{makecell}
\usepackage{stfloats}
\usepackage{graphicx}
\usepackage{textcomp}
\usepackage{xcolor}
\usepackage{pifont}
\usepackage{algorithm}  
\usepackage[normalem]{ulem}  
\usepackage{algpseudocode}   
\usepackage{multirow}
\usepackage{fix-cm}
\renewcommand\footnotetextcopyrightpermission[1]{} 
\begin{document}


\title
{\vspace{-0.5cm} CorrectHDL: Agentic HDL Design with LLMs Leveraging High-Level Synthesis as Reference\vspace{-0.15cm}}

\author{Kangwei Xu\textsuperscript{1}, Grace Li Zhang\textsuperscript{2}, Ulf Schlichtmann\textsuperscript{1}, Bing Li\textsuperscript{3}\\}
\affiliation{%
  \textsuperscript{1}\textit{ Technical University of Munich (TUM)}, Munich\country{Germany}\\
\textsuperscript{2}\textit{Technical University of Darmstadt}, Darmstadt\country{Germany}\\
\textsuperscript{3}\textit{Technical University of Ilmenau}, Ilmenau\country{Germany}\vspace{-0.1cm} }

\renewcommand{\shortauthors}{}

\begin{abstract} 
Large Language Models (LLMs) have demonstrated remarkable potential in hardware front-end design using hardware description languages (HDLs). However, their inherent tendency toward hallucination often introduces functional errors into the generated HDL designs. To address this issue, we propose the framework \textit{CorrectHDL} that leverages high-level synthesis (HLS) results as functional references to correct potential errors in LLM-generated HDL designs.
The input to the proposed framework is a C/C++ program that specifies the target circuit’s functionality. 
The program is provided to an LLM to directly generate an HDL design, whose syntax errors are repaired using a Retrieval-Augmented Generation (RAG) mechanism. The functional correctness of the LLM-generated circuit is iteratively improved by comparing its simulated behavior with an HLS reference design produced by conventional HLS tools, which ensures the functional correctness of the result but can lead to suboptimal area and power efficiency. 
Experimental results demonstrate that circuits generated by the proposed framework achieve significantly better area and power efficiency than conventional HLS designs and approach the quality of human-engineered circuits. Meanwhile, the correctness of the resulting HDL implementation is maintained, highlighting the effectiveness and potential of agentic HDL design leveraging the generative capabilities of LLMs and the rigor of traditional correctness-driven IC design flows. This work is open-sourced at the following link: \underline{\url{https://github.com/AgenticHDL/CorrectHDL}}
\end{abstract}

\maketitle
\pagestyle{empty}
\fancyhead{} 

\section{Introduction}
\begin{figure}[t]
\vspace{-0.7cm}
\centering	\includegraphics[width=1\linewidth]{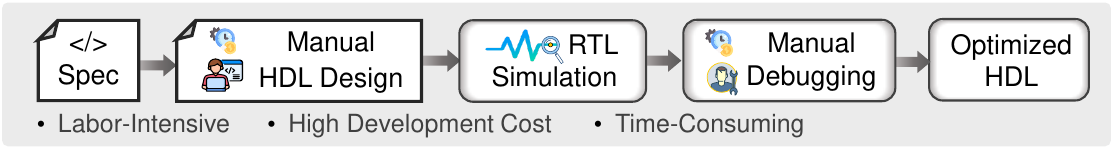}
	\vspace{-0.75cm}
	\caption{~Traditional manual HDL design flow.}
	\label{fig:trad}
	\vspace{-0.3cm}
\end{figure}

With the growing complexity of integrated circuits, hardware engineers are required to invest increasing effort in HDL design \cite{b1}. Traditional hardware design flows typically begin with engineers interpreting algorithm specifications and then manually 
coding them into corresponding HDL designs. As shown in Fig.~\ref{fig:trad}(a), this process is highly labor-intensive: developers should write HDL designs by hand, perform simulation, and debug errors to ensure functional correctness, resulting in a time-consuming and costly process \cite{b2}.

Recently, large language models (LLMs) have demonstrated significant potential in enhancing the efficiency of HDL design \cite{b6.1, b6.2, b6.3, b6.4, b6.5, b6.6, b6.7, b6.8, b6.9, b6.10, b6.11, b6.12, b6.13, b6.14, b6.15}. By learning expert design patterns, LLMs can generate HDL designs directly from specifications with significantly reduced manual effort \cite{b7.1, b7.2, b7.3, b7.4, b7.5, b7.7, b7.8, b7.9, b7.10}. Chip-Chat \cite{b6.1} presents the first systematic study of conversational LLMs for interactively co-designing an 8-bit accumulator microprocessor under real-world constraints. Rome \cite{b6.2} introduces hierarchical prompting and an automated generation pipeline that enables LLMs to reliably produce multi-level HDL designs. GPT4AIGChip \cite{b6.3} leverages LLMs for AI accelerator design by decoupling circuit modules with in-context learning. RTLFixer \cite{b6.4} introduces a RAG-based debugging flow that repairs syntax errors in HDL designs. 
AssertLLM \cite{b8.4} generates functional assertions from specifications, while \textit{CorrectBench} \cite{b8.5, b8.6} employs LLMs to build a hybrid evaluation platform with a self-correction mechanism. The frontier of LLM-assisted hardware design is further advanced in High-Level Synthesis (HLS) \cite{b10.0, b10.1, b10.11, b10.2, b10.3, b10.4, b10.5}, C2HLSC \cite{b10.0, b10.1, b10.11}, HLS-Repair \cite{b10.2}, and HLSPilot \cite{b10.3} use LLMs to iteratively refactor C/C++ programs into HLS-compatible versions with tool-guided feedback. 

Despite the advances of HDL design with LLMs, the inherent hallucinations of LLMs frequently introduce syntactic and, more critically, functional errors into the generated HDL designs \cite{b6.1}. While syntax issues can often be detected by compilers and fixed by iterative repair loops \cite{b6.4}, functional debugging remains fundamentally challenging. Current LLM-based flows lack a bit-accurate reference at the same abstraction level as the generated HDL, forcing engineers to manually interpret algorithm specifications, analyze simulation waveforms, and infer intended behaviors to locate functional bugs. Consequently, the effort required to validate and correct LLM-generated designs can partially offset the benefits of automated code generation, thereby limiting the practical adoption of LLMs for producing functionally correct HDL in industry \cite{b7.4, b7.5}. 

To address the challenge above, we introduce a novel agentic design framework, \textit{CorrectHDL}, for LLM-assisted HDL design and debugging using the results of high-level synthesis (HLS) as a bit-accurate reference to overcome functional errors. To the best of our knowledge, this is the first work leveraging the generative capabilities of LLMs and the rigor of traditional correctness-driven synthesis flows to support LLM-assisted HDL design and debugging. The key contributions of this paper are summarized as follows:

\begin{itemize}
\item 
An agentic HDL design framework, \textit{CorrectHDL}, is proposed and released as open source. By leveraging HLS-generated HDL as a functional reference, \textit{CorrectHDL} implements an end-to-end design flow that covers HDL generation, runtime profiling, differential verification, iterative repair, and PPA evaluation, enabling accurate and efficient hardware design.

\item 
To mitigate attention dilution in LLMs, complex C/C++ algorithms are decomposed into LLM-friendly submodules through a structured C/C++ decomposition strategy, which also provides a basis for subsequent differential debugging.

\item 
By leveraging the adaptive learning capability of LLMs, a Retrieval-Augmented Generation mechanism is introduced to correct syntax errors in the LLM-generated HDL designs, increasing the compilation pass rate by an average of 15.49\%.

\item 
To address functional errors caused by LLM hallucinations, a differential verification and repair loop is established using HLS-generated HDL as the functional reference, improving the functional pass rate of LLM-generated HDL by 28.05\%.  

\item 
To efficiently debug the LLM-generated top-level HDL design, submodule boundary instrumentation is combined with a backward slicing technique to locate errors from the system level down to individual modules, thereby bringing the overall functional pass rate improvement to 38.54\%. Experimental results also demonstrate that the proposed \textit{CorrectHDL} achieves an average area reduction of 24.83\% and power reduction of 26.98\% compared with conventional HLS designs and approaches the quality of human-engineered circuits.

\end{itemize}

The paper is organized as follows. Section~\ref{sec:1second} provides the background and motivation. Section~\ref{sec:1third} details the framework. Section~\ref{sec:fourth} shows the experimental results, and Section~\ref{sec:fifth} concludes the paper.

\begin{figure}[t]
\vspace{-0cm}
\centering	\includegraphics[width=1\linewidth]{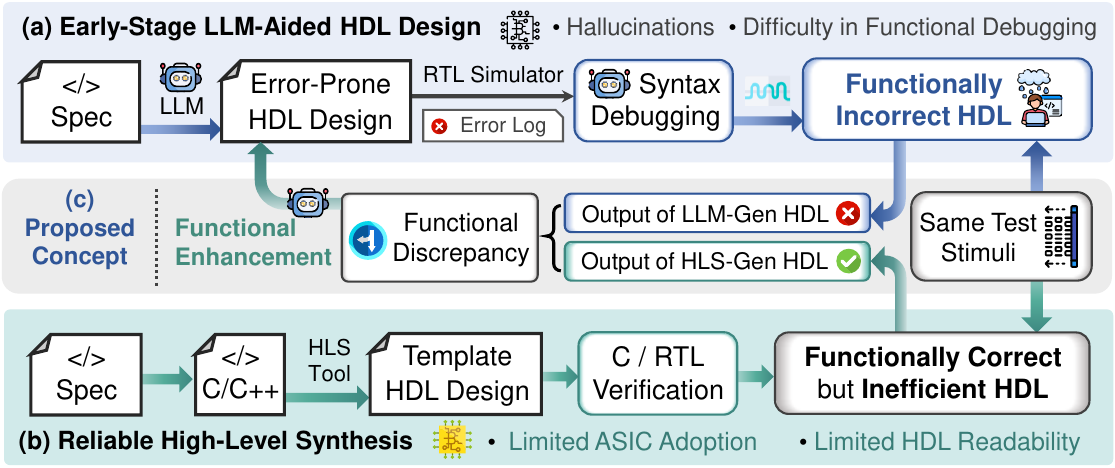}
	\vspace{-0.7cm}
	\caption{~(a) LLM-aided HDL design flow; (b) High-Level Synthesis (HLS) flow; (c) Our proposed concept.}
	\label{fig:back}
	\vspace{-0.5cm}
\end{figure}

\vspace{0.15cm}
\section{Motivation and Concept} \label{sec:1second}
To address the functional errors in the HDL design generated by LLMs, an effective approach is to use a reference HDL design to guide the code generation process of LLMs. Such a functional reference should be able to establish the mapping from a high-level description to HDL that accurately reflects the intended circuit behavior. The performance, power, and area (PPA) of such a reference are not critical, because only the HDL generated by LLMs is used in real circuit design. To create such a reference, the appropriate format of the high-level description and a verified toolchain that can generate an HDL design from this description should be identified.

Over the past decades, the EDA community has proposed a wide range of solutions for circuit design based on abstract behavioral or algorithmic descriptions, among which high-level synthesis (HLS) has emerged as the most prominent approach for translating C/C++ programs into HDL designs \cite{b2, b3}. Modern HLS toolchains are mature and highly reliable, achieving notable success in both industry and academia \cite{b4, b5}. Unlike direct software-level simulation, the HLS-generated circuit reference preserves precise hardware semantics, accurately handling customized bit widths, type conversions, bit truncation/rounding modes. This eliminates the semantic ambiguities inherent in software simulation. However, the underlying compilation flows of HLS tools typically rely on fixed templates and coarse-grained pragma tuning, leading to suboptimal area and power efficiency of the synthesized circuit \cite{b24, b24.1}. In addition, the HDL designs generated by HLS are often difficult to interpret, which makes late-stage Engineering Change Orders (ECOs) challenging in ASIC-oriented physical design \cite{b24.2}. Consequently, for high-performance designs, engineers still prefer to manually craft HDL based on the algorithm descriptions to meet stringent PPA requirements \cite{b25}.

Given the ability of HLS to generate functionally correct HDL designs, our agentic design framework, \textit{CorrectHDL}, takes advantage of the results of HLS to guide LLMs in automatic HDL generation. In this framework, the generative capabilities of LLMs are leveraged to produce flexible HDL, while the HLS tool provides a bit-accurate golden reference to guide the functional correctness of LLM-generated HDL. Since LLMs do not use predefined circuit templates but rely on abstracted design knowledge, they can flexibly create circuits that outperform HLS-generated designs in terms of PPA and approach the quality of human-engineered circuits.

\begin{figure}[t]
\vspace{-0.7cm}
\centering	
\includegraphics[width=0.96\linewidth]{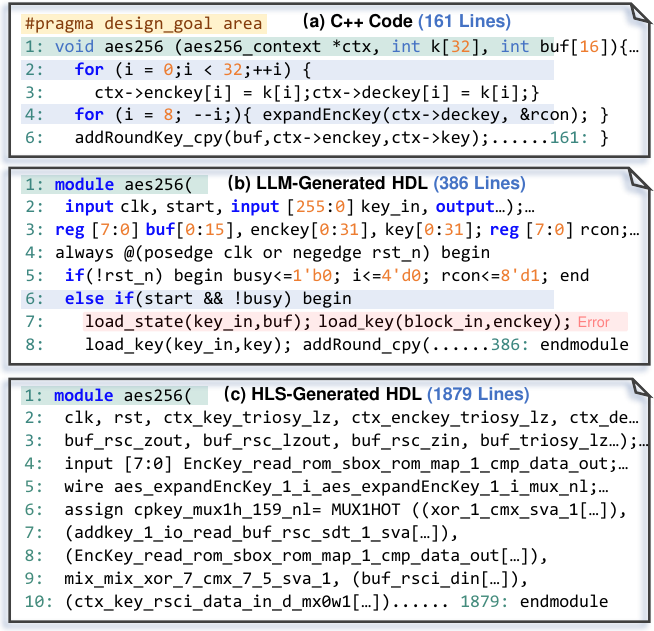}
\vspace{-0.4cm}
\caption{Examples of HDL Generation via LLM and HLS.}
\label{fig:code}
\vspace{-0.63cm}
\end{figure}

The concept of \textit{CorrectHDL} is illustrated in Fig.~\ref{fig:back}. The circuit specification written in C/C++ is first processed by an HLS tool, which produces a functionally correct HDL design that serves as the golden reference. The functional behavior of the HDL generated by the LLM is then compared with that of the reference design from HLS. The discrepancy is used to guide the LLM to iteratively enhance the HDL designs to achieve functional correctness.

The input to the \textit{CorrectHDL} framework consists of circuit descriptions expressed in C/C++, which provide a more detailed specification than the natural language inputs used in many state-of-the-art solutions \cite{b7.1, b7.2, b7.3, b7.4, b7.5, b7.7, b7.8, b7.9, b7.10}. This choice makes the framework directly applicable to system design engineers who typically work at the algorithm level. Meanwhile, natural language specifications can be automatically translated into C/C++ using LLMs, which have demonstrated high accuracy in code generation within the software community~\cite{b17, b18}. Accordingly, the proposed \textit{CorrectHDL} effectively bridges an important gap in LLM-based circuit design.

The output of the \textit{CorrectHDL} framework is a circuit generated by the LLM. 
Its functionality is iteratively improved by comparing its behavior against a reference HDL produced by HLS. The correctness of this reference HDL is validated using a testbench that the HLS tool automatically derives from the original C/C++ testbench. However, this converted testbench is coupled to the tool-specific interfaces of the synthesized design, so it cannot be directly used to simulate the circuit generated by the LLM agent. To address this issue, we also employ LLMs to translate the original C/C++ testbench into an HDL testbench that is compatible with the LLM-generated HDL. Limited human effort is still required during this process to ensure the quality of the resulting testbench. Functional comparison is then performed by evaluating the output data of the LLM-generated circuit against that of the HLS reference under the same test stimuli.

Fig.~\ref{fig:code}(b) illustrates an example of an LLM-generated HDL design from the C/C++ program in Fig.~\ref{fig:code}(a), which exhibits good readability. However, limited hardware-oriented training data and LLM hallucinations often lead to functional errors (line 7). On the contrary, the code in Fig.~\ref{fig:code}(c) generated by HLS is functionally correct, but it has far inferior readability and PPA. The objective of \textit{CorrectHDL} is thus to enhance the functional correctness of the LLM-generated HDL design using the HLS-generated HDL design as a reference.

\begin{figure}[t]
\vspace{-0.6cm}
\centering	\includegraphics[width=0.99\linewidth]{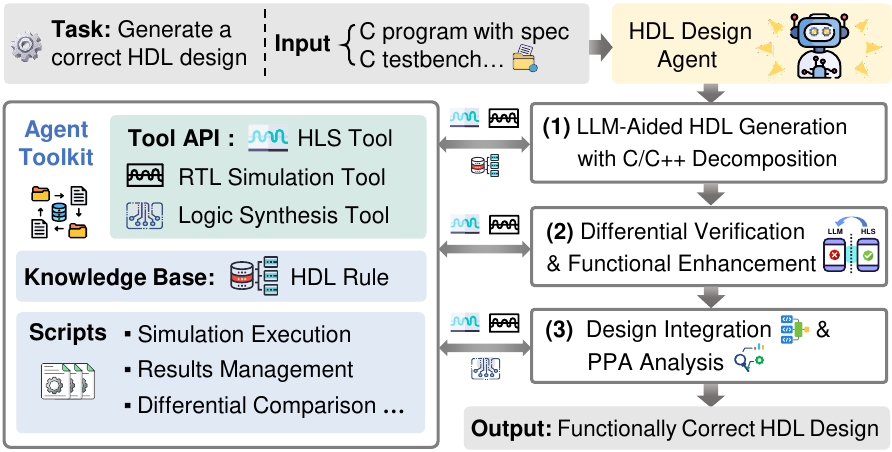}
	\vspace{-0.35cm}
	\caption{~Proposed HDL design agent.}
	\label{fig:agent}
	\vspace{-0.3cm}
\end{figure}


\begin{figure}[t]
\vspace{-0cm}
\centering	
\includegraphics[width=0.99\linewidth]{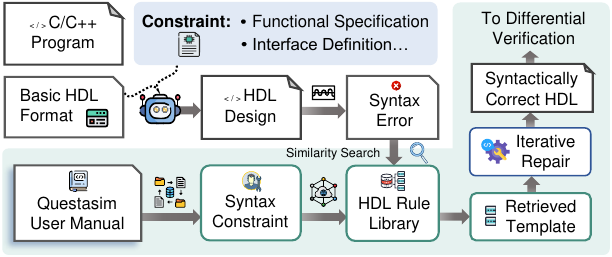}
\vspace{-0.35cm}
\caption{~LLM-assisted HDL generation.}
\label{fig:hdlgen}
\vspace{-0.4cm}
\end{figure}

\section{Agentic HDL Design with LLMs Leveraging HLS as Reference}\label{sec:1third}

The overall workflow of the proposed HDL Design Agent is shown in Fig.~\ref{fig:agent}. Given a C/C++ program with its specification and a corresponding C/C++ testbench from the HLS benchmark suite \cite{b40, b41}, the agent executes a three-stage process to produce a functionally correct HDL design. The agent interacts with a toolkit comprising tool APIs (HLS, RTL simulation, logic synthesis), a knowledge base of HDL syntax rules, and automation scripts for simulation and comparison. In stage (1), described in Section 3.1, the C/C++ program is decomposed into submodules according to the rules in Section~\ref{sec:first1}. For each submodule, the LLM generates a corresponding HDL module (Design Under Test, DUT), and syntax errors are repaired through the RAG mechanism. In stage (2), described in Section 3.2, each C/C++ submodule is synthesized by the HLS tool into a functionally correct HDL as the golden reference. The DUT and the golden reference are then simulated under the same stimuli. During this process, HLS tools can automatically translate the original C/C++-based testbench into an equivalent HDL testbench to validate the HLS-generated HDL design, whereas the testbench for the LLM-generated HDL design is adapted from the original C/C++ testbench using the LLM and then verified by humans. Any detected functional mismatches are fed back to the LLM for iterative enhancement. In stage (3), described in Section 3.3, all verified LLM-generated submodules are instantiated into a top-level design, and differential verification is repeated to ensure integration correctness.

\vspace{-0.2cm}
\subsection{LLM-Assisted HDL Design Generation}\label{sec:first}
\subsubsection{C/C++ Program Decomposition}\label{sec:first1}
Directly feeding an entire C/C++ program with long context into the LLM tends to dilute attention, preventing it from capturing fine-grained hardware semantics, e.g., customized bit widths, type conversions, etc. This often leads to deep functional errors that are difficult to debug.
To mitigate this issue, a divide-and-conquer strategy based on C/C++ decomposition is adopted before HDL generation. The original C/C++ program is decomposed into smaller, LLM-friendly C/C++ submodules, thereby improving the quality of the initial HDL design generation. 

The decomposition follows three strict rules: (1) Function-Level Granularity: Decomposition is performed along function boundaries, rather than by arbitrary line-level splitting. (2) Explicit I/O definition: Each submodule uses fixed-width scalars or static arrays as interfaces (e.g., int8\_t state[16]), with all bit widths explicitly specified. (3) Single and Clear Semantics: Each submodule should have clear semantics (e.g., SubBytes() performs only SBox substitution).

To ensure correctness after decomposition, a differential verification step is performed. All decomposed C/C++ submodules are re-integrated into a top-level C/C++ program and verified against the original C/C++ testbench. The decomposition is considered successful only when the output results match exactly.

\begin{figure}[t]
\vspace{-0.55cm}
\centering	
\includegraphics[width=1.01\linewidth]{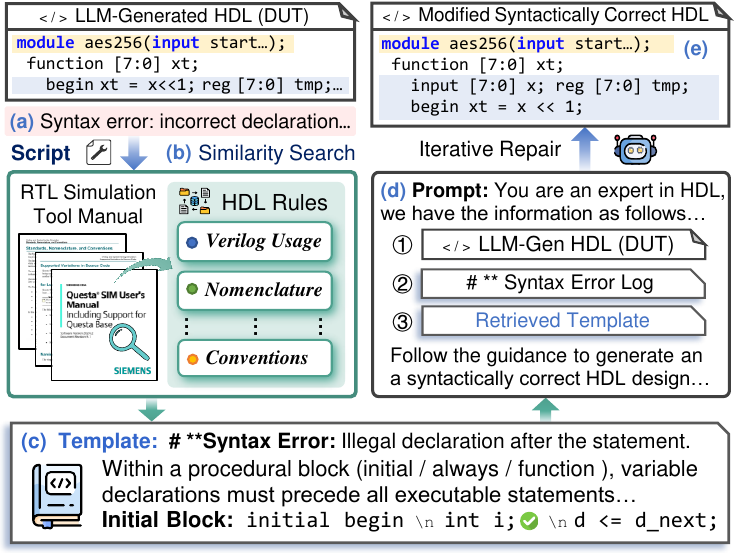}
\vspace{-0.75cm}
\caption{~Syntax repair for LLM-generated HDL using RAG.}
\label{fig:rag}
\vspace{-0.6cm}
\end{figure}

\vspace{-0.2cm}
\subsubsection{LLM-Assisted HDL Design Generation}\label{sec:first2}

After decomposition, the agent proceeds to the HDL generation phase as shown in Fig.~\ref{fig:hdlgen}.
For each C/C++ submodule, we generate two specification files in natural language using the LLM. (1) Functional Specification: A detailed description of the functionality and expected hardware behavior of the C/C++ submodule. (2) Interface Definition: Precisely maps C/C++ parameters to Verilog ports, including names, directions, and exact bit widths. With these two specification files, the LLM can better understand the design constraints of individual submodules and produce valid HDL designs from the C/C++ snippet effectively.

To guide the LLM toward producing effective HDL designs, we also define the design, and formatting constraints as follows: 
(1) Separating control logic from the datapath and using synchronous reset conventions; (2) Optimization strategies for the circuits, e.g., inserting pipeline stages; (3) Formatting constraints such as enclosing the resulting HDL between triple backticks to facilitate HDL extraction from the output of the LLM using a script. 
The C/C++ submodule code, functional specification, interface definition, design and formatting constraints are jointly provided to the LLM, guiding it to generate the corresponding HDL design of a submodule.

Due to LLM hallucinations, the HDL design generated for a submodule often fails syntax compilation. As shown in Fig.~\ref{fig:hdlgen}, these errors can be repaired by a retrieval-augmented generation (RAG) mechanism, which involves three stages: First, an external rule library is created containing HDL syntax rule templates. Each template includes error logs that may be triggered during syntax compilation, and a summary of the corresponding repair rule extracted from the user manual. Second, when an LLM-generated HDL design fails compilation, the error log is embedded and then matched against the templates in a rule library using a sentence transformer \cite{b38}. 
The retrieval process can be formalized as follows:
\vspace{-0.1cm}
\begin{equation}\label{eq:define}
E = \mathrm{embed}(\text{error\_log}), \quad
R_i = \mathrm{embed}(\text{rule\_template}_i),
\vspace{-0.1cm}
\end{equation}
where \(E\) and \(R_i\) denote the embedding vectors of the error log and the \(i\)th rule template. Given an error log $E$ and a template $R$, the rule template with the highest similarity is selected as follows:
\vspace{-0.2cm}
\begin{equation}\label{eq:cosine_selection}
\mathrm{sim}(E,R) = \frac{E \cdot R}{\|E\|\,\|R\|}~, \quad
R^{*} = \underset{R_i \in \mathcal{R}}{\operatorname{arg\,max}}\, \mathrm{sim}(E,R_i),
\vspace{-0.2cm}
\end{equation}
where $\mathcal{R}$ denotes all templates in the library, $R^{*}$ represents the template with the highest similarity. Third, once the most relevant rule template is retrieved, it is incorporated into the LLM prompt to guide the repair of the HDL design into a syntactically correct version. An example of repairing an LLM-generated HDL design is given in Fig.~\ref{fig:rag}. The agent uses compiler logs to retrieve repair rules and then prompts the LLM to repair the HDL, and repeats this process until compilation succeeds or the predefined iteration limit is reached.

\begin{figure}[t]
\vspace{-0.2cm}
\centering	
\includegraphics[width=1\linewidth]{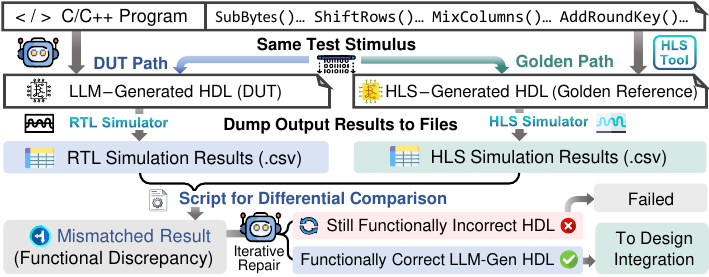}
\vspace{-0.7cm}
\caption{~Differential verification for LLM-generated HDL.}
\label{fig:veri}
\vspace{-0.3cm}
\end{figure}

\begin{figure}[t]
\vspace{-0cm}
\centering
	\includegraphics[width=1\linewidth]{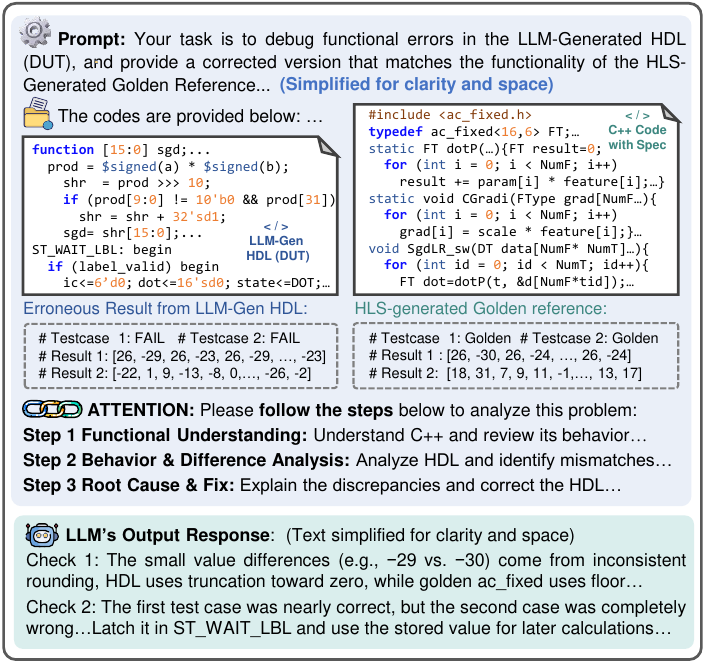}
	\vspace{-0.7cm}
	\caption{Functional enhancement in LLM-generated HDL.}
	\label{fig:funcre}
	\vspace{-0.6cm}
\end{figure}

\vspace{-0cm}
\subsection{Differential Verification and Functional Enhancement}\label{sec:second}
After syntax errors in LLM-generated HDL designs are corrected as described in Section~\ref{sec:first2}, we leverage the HLS-generated HDL as a golden functional reference to enhance its functional correctness, as shown in Fig.~\ref{fig:veri}. For each C/C++ submodule, the HLS tool synthesizes a bit-accurate and functionally correct HDL design as a reference. Despite its suboptimal area/power efficiency and limited readability, its functional behavior is highly reliable \cite{b2, b3, b4, b5} and thus serves as an effective golden reference for verifying LLM-generated design.

To enable bit-accurate comparison, the LLM-generated HDL design as DUT and the HLS-generated HDL design as golden reference are simulated under identical test stimuli derived from the C/C++ testbench. Modern HLS tools can automatically translate the C/C++ testbench into an equivalent HDL testbench to test the HLS-generated HDL. The corresponding HDL testbench for the LLM-generated HDL is adapted from the same C/C++ testbench via the LLM with human verification to ensure its correctness in validating the LLM-generated HDL design. The new testbench is examined by human engineers to ensure its quality.

\begin{table}[!t]
\vspace{-0.2cm}
\centering	
  \refstepcounter{table}%
  {TABLE 1: Typical Functional Discrepancies in LLM-Generated HDL and the Corresponding Corrections Driven by \textit{CorrectHDL}.\par}
\vspace{0.1cm}
\includegraphics[width=1.01\linewidth]{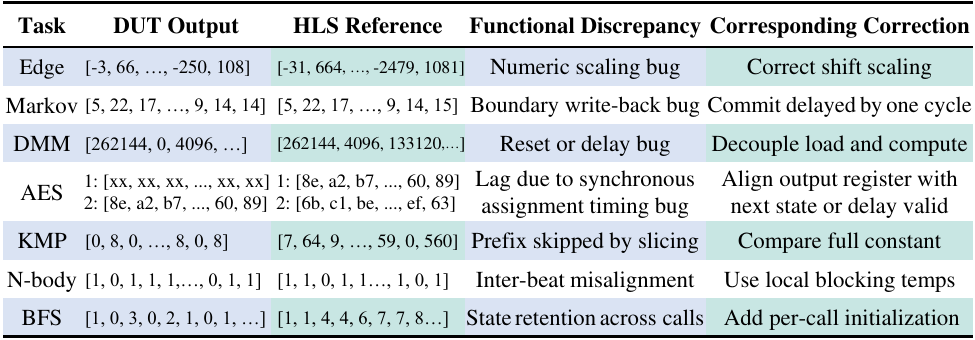}
	\label{tab:discre}
	\vspace{-0.8cm}
\end{table}

The agent then initiates a parallel simulation flow shown in Fig.~\ref{fig:veri}, where in the DUT Path, the LLM-generated HDL design is simulated in an RTL simulation environment, and in the Golden Path, the HLS-generated HDL design is simulated in the HLS simulation environment.
During both simulations, all output ports are recorded and dumped into result files. A comparison script conducts automated differential checking between these files.  If all outputs match, the DUT is considered functionally correct and collected for subsequent integration.  If any mismatch is observed, the DUT is marked as functionally incorrect, and the functional repair loop is triggered.

The repair loop uses the mismatched results as precise feedback to guide the LLM in the functional repair of the DUT. A structured prompt is constructed that contains the current (incorrect) Verilog code of the DUT, the C/C++ program that defines the intended functional behavior, and a mismatch log produced by differential comparison. This mismatch log explicitly records the test stimulus, the actual DUT output, and the expected golden reference for each failing test case.
In addition, a step-wise reasoning method inspired by Chain-of-Thought is adopted to repair the LLM-generated HDL design. As shown in Fig.~\ref{fig:funcre}, the prompt explicitly instructs the LLM to strengthen functional understanding by analyzing the C/C++ program and summarizing the intended behavior. It then conducts behavior and difference analysis through reviewing the HDL design, correlating it with the expected functionalities, and identifying the source of the mismatches. The root cause of errors is identified and fixed by explaining the simulation discrepancy. With such concrete feedback, the LLM can then perform targeted debugging rather than blind editing. If mismatches remain, the verification–feedback–repair iteration is repeated until functional equivalence is achieved or the predefined iteration limit is reached.

\begin{table*}[t]
\vspace{-0.6cm}
\centering
  \refstepcounter{table}%
  {TABLE 2: Comparison of \textit{CorrectHDL} with Baselines in Simulation Pass Rate, and with HLS in Area and Power Efficiency.\par}
	\vspace{0.05cm}
	\includegraphics[width=0.97\linewidth]{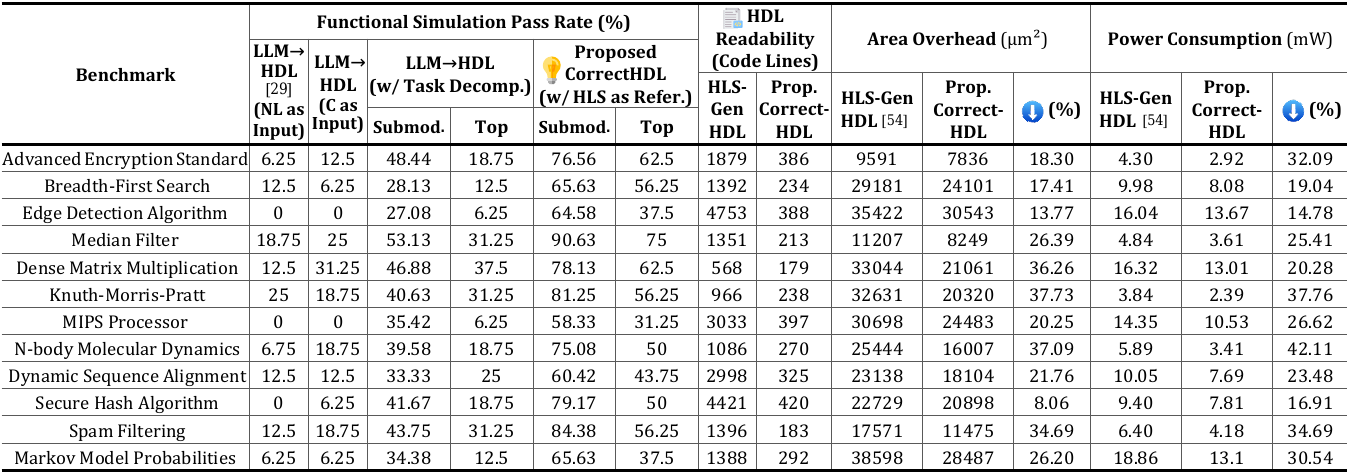}
	\label{tab:table}
	\vspace{-0cm}
	\begin{tablenotes}
\item[] \scriptsize * The simulation pass rate is calculated based on 16 independent generation rounds by the LLM for each task.
\end{tablenotes}
\vspace{-0.55cm}
\end{table*}

A concrete example of the LLM-driven reasoning chain is shown in Fig.~\ref{fig:funcre}. Guided by this structured prompt, the LLM first identifies the small value differences stemming from inconsistent rounding semantics between the DUT and the golden fixed-point implementation. It then proposes a revised HDL version that corrects the rounding behavior. In the second check, the LLM observes that the first test case is almost correct but the second is completely wrong, and traces this to a missing state retention in the control logic. It proposes latching the intermediate result in the \texttt{ST\_WAIT\_LBL} state and reusing the stored value in subsequent calculations. After re-simulation, the repaired DUT matches the golden outputs on all test cases, demonstrating how the agent can systematically analyze and repair functional discrepancies with reliable functional references.
\begin{figure}[t]
\vspace{0.2cm}
\centering
	\includegraphics[width=1\linewidth]{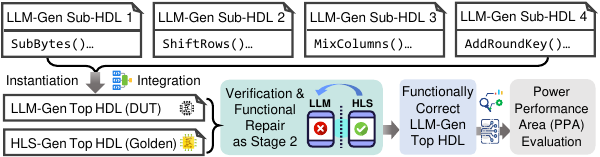}
	\vspace{-0.73cm}
	\caption{Design integration.}
	\label{fig:deinter}
	\vspace{-0.7cm}
\end{figure}
To better describe the types of functional errors handled by our framework, Table~\ref{tab:discre} summarizes several typical functional discrepancies observed between the DUT and the golden reference. For each task, the table lists representative DUT and HLS golden outputs and highlights how the LLM identifies the underlying root cause, such as numeric scaling errors, reset or lag issues, or state-retention faults, and generates the corresponding fixes, including correcting fixed-point scaling, decoupling load and compute operations, or adding per-call initialization. These examples indicate that the encountered bugs are not simple syntactic issues but functional errors that are difficult to debug without a golden functional reference, highlighting its essential role in the repair process.

\vspace{-0.2cm}
\subsection{Design Integration}\label{sec:third}
After the LLM-generated HDL designs of all submodules pass differential verification, they are integrated into a functionally correct top-level HDL design. To enable correct instantiation and wiring, the LLM is prompted with the original C/C++ program from Section~\ref{sec:first1} and all verified HDL designs from Section~\ref{sec:second} together with their interface definitions from Section~\ref{sec:first2}. Using this information, function calls and variable transfers in the C/C++ code are systematically mapped to HDL module instances and signal connections, ensuring that data dependencies and interface contracts are preserved. The result is an LLM-generated top-level HDL design that implements the behavior of the original C/C++ program.

As shown in Fig.~\ref{fig:deinter}, the integrated top-level HDL design may still contain errors that are not exposed at the submodule level, such as misconnected ports or misaligned valid signals. 
To validate the top-level behavior, the original C/C++ program is then synthesized by the HLS tool into an HLS-generated top HDL design, which serves as the golden reference. Differential verification is then performed at the top level, as in Section~\ref{sec:second}, to detect integration errors.

To efficiently locate integration errors, boundary signal instrumentation is inserted at all submodule interfaces in the top-level design. During verification, the agent compares not only the final outputs but also these internal boundary signals against the golden reference. Guided by this structured feedback, the LLM performs targeted repair by using the backward slicing technique to trace the origin of the discrepancy \cite{b39}. The repair process is repeated until functional equivalence is achieved or the iteration limit is reached.

\vspace{-0.2cm}
\section{Experimental Results}\label{sec:fourth}
To evaluate the quality of HDL designs generated by the proposed \textit{CorrectHDL}, we applied it to 12 real-world tasks~\cite{b40, b41} and measured the syntax compilation and functional simulation pass rate of the generated HDL designs. We also evaluated the area and power efficiency of the HDL designs produced by \textit{CorrectHDL} and compared them with those generated by traditional HLS tools~\cite{b41.1} and manually written designs from open-source implementations~\cite{b42, b43}.

In the evaluation, the Catapult HLS tool and QuestaSim RTL simulator \cite{b41.1, b41.2} were used to generate and simulate the HDL designs. The area and power of the HDL designs were evaluated by synthesizing them using Design Compiler with the Nangate 45nm library. In \textit{CorrectHDL}, the GPT-5 model was employed as the LLM via OpenAI APIs \cite{b26}. 
To ensure reliability and robustness, each task was repeated $n$ times. $n$ = 16 was used in the experiments. For each time, the LLM was queried three times to iteratively correct the HDL based on tool feedback. The pass rate is computed as $\text{Pass Rate (\%)}= m/n$, where $m$ represents the number of successfully generated HDL designs and $n$ denotes the total number of HDL designs.

Table~\ref{tab:table} compares the HDL designs generated by \textit{CorrectHDL} with those directly generated using the LLM with C/C++ codes as inputs and with HLS-generated HDL designs. Column 1 lists the benchmark names. Column 2 reports the simulation pass rate of directly prompting the LLM with natural language~\cite{b7.10}. This setting leads to low simulation pass rates, indicating that most generated HDL designs exhibit functional errors. Column 3 shows that prompting the LLM with the C/C++ program still results in low simulation pass rates, although it performs better on average than natural language prompts because C/C++ more precisely captures the underlying algorithmic behavior. With the C/C++ decomposition step in \textit{CorrectHDL}, the simulation pass rate can be enhanced, as shown in columns 4-5, where column 4 reports the average pass rate of decomposed submodules and column 5 shows the pass rate of the top design by integrating such submodules together by the LLM.
Columns 6-7 show the functional simulation rate of submodules and top design of \textit{CorrectHDL}, which demonstrates significant improvements in functional correctness of the LLM-generated HDL.

\begin{figure}[t]
\vspace{-0.5cm}
\centering	
\includegraphics[width=1\linewidth]{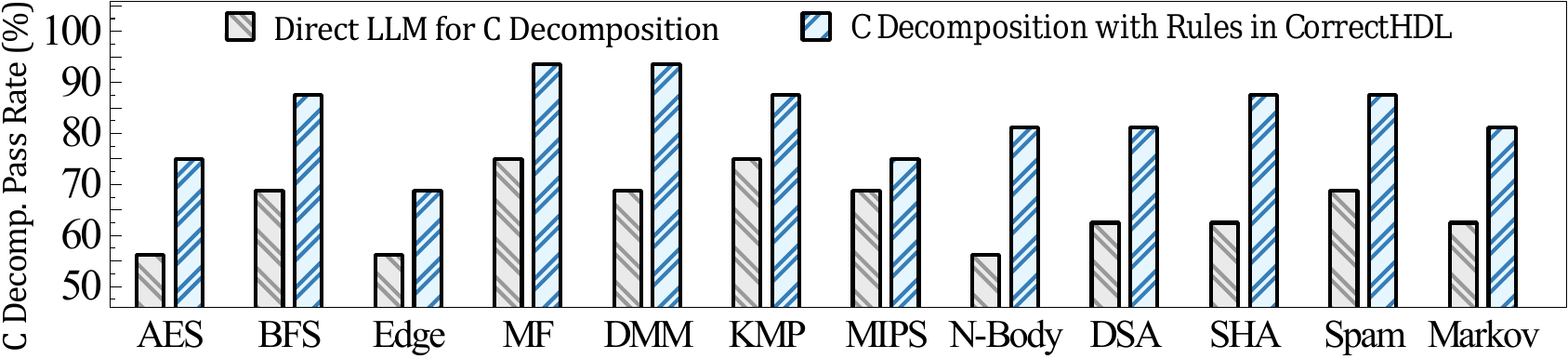}
\vspace{-0.8cm}
\caption{\fontsize{8.8pt}{9pt}\selectfont C/C++ decomposition pass rate.}
\label{fig:cdecom}
\vspace{-0.35cm}
\end{figure}

\begin{figure}[t]
\vspace{-0cm}
\centering	
\includegraphics[width=1\linewidth]{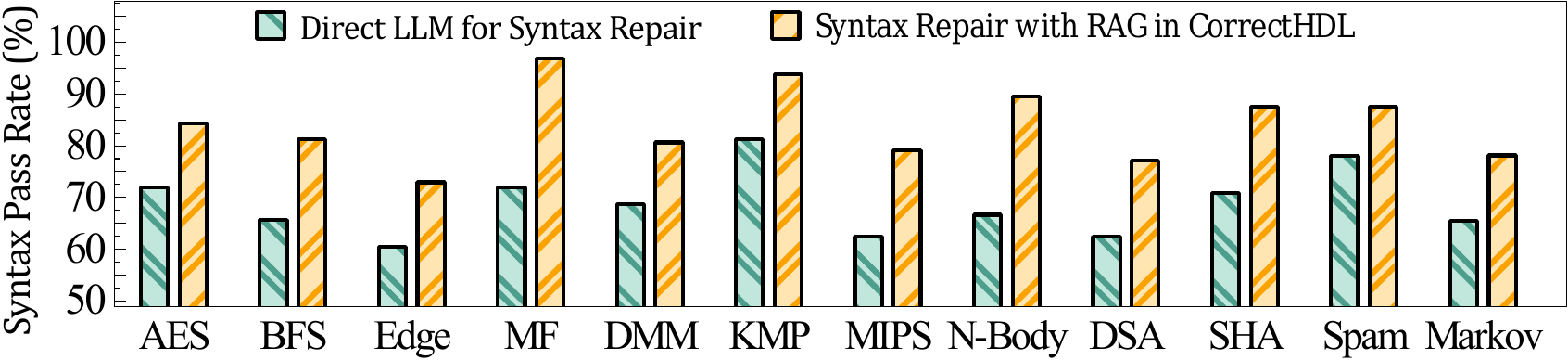}
\vspace{-0.8cm}
\caption{\fontsize{8.8pt}{9pt}\selectfont Syntax pass rate of the LLM-generated HDL.}
\label{fig:ragre}
\vspace{-0.35cm}
\end{figure}

\begin{figure}[t]
\vspace{-0cm}
\centering	
\includegraphics[width=1\linewidth]{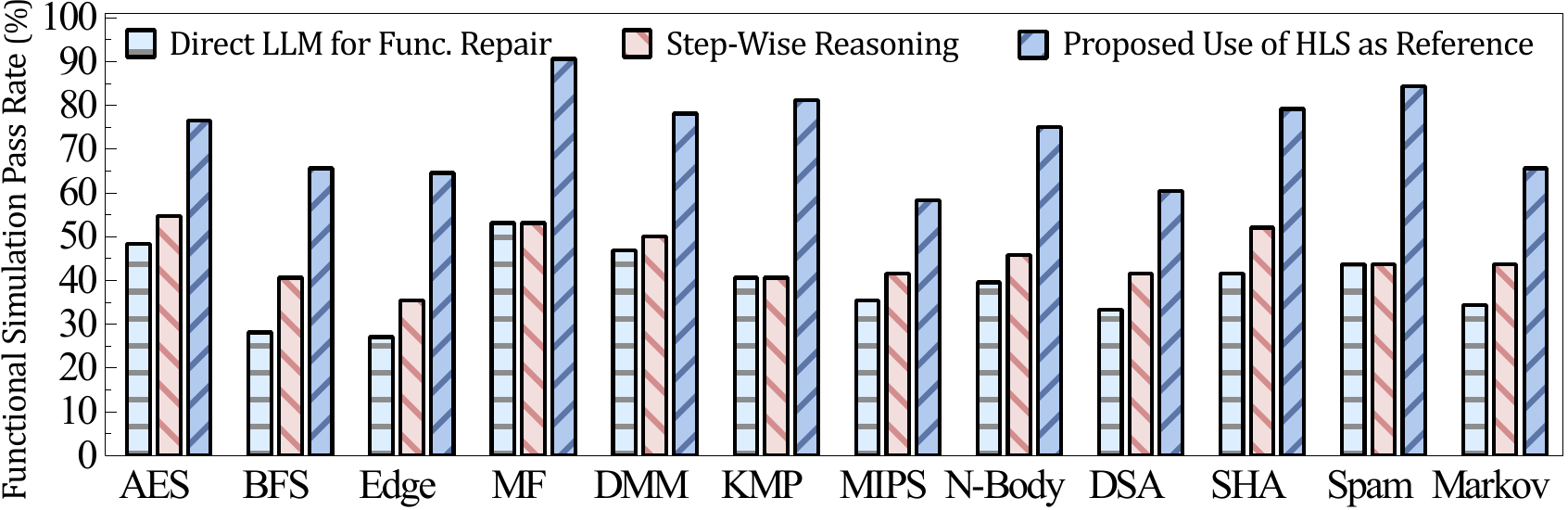}
\vspace{-0.8cm}
\caption{\fontsize{8.8pt}{9pt}\selectfont Functional pass rate of the LLM-generated sub-HDL.}
\label{fig:funcpr}
\vspace{-0.35cm}
\end{figure}

\begin{figure}[t]
\vspace{-0cm}
\centering	
\includegraphics[width=1\linewidth]{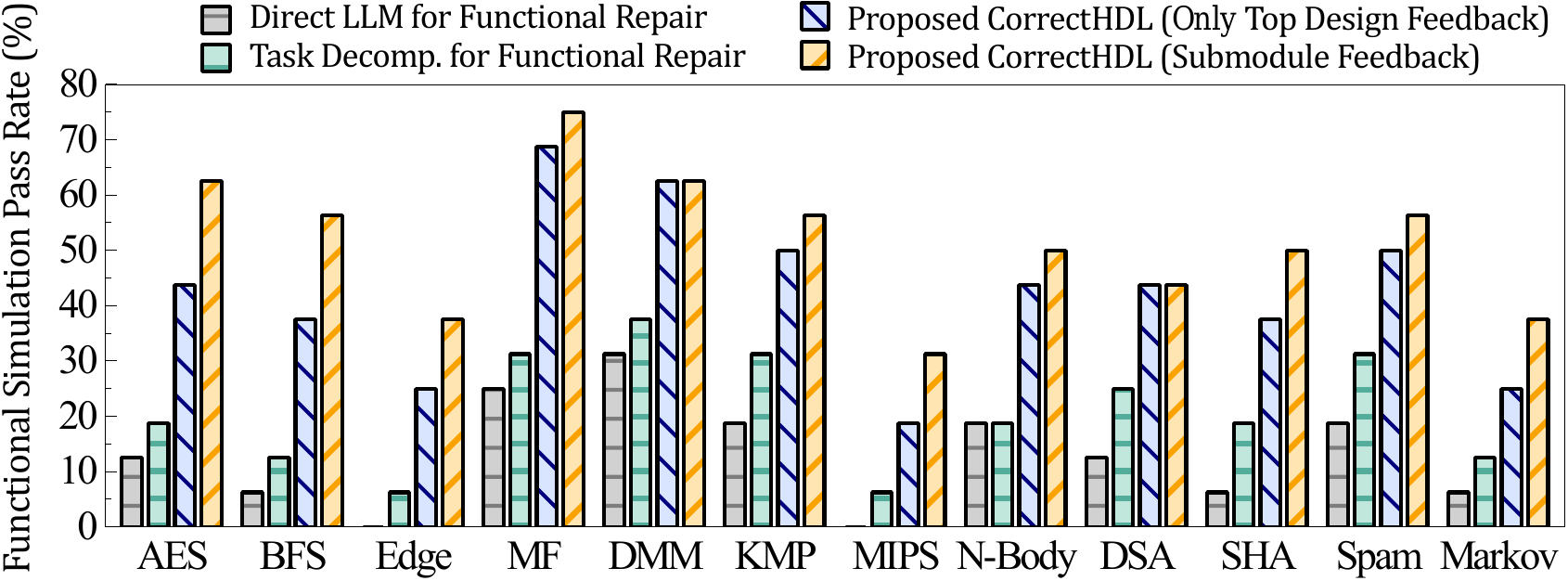}
\vspace{-0.8cm}
\caption{\fontsize{8.8pt}{9pt}\selectfont Functional pass rate of the LLM-generated top HDL.}
\label{fig:interpr}
\vspace{-0.5cm}
\end{figure}

To compare the quality of HDL code generated by \textit{CorrectHDL} and the traditional HLS tool, 
columns 8–9 in Table~\ref{tab:table} report readability and maintainability of HDL code, measured by the number of lines of HDL code. 
This comparison shows \textit{CorrectHDL} consistently generates more readable HDL than HLS. 
Columns 10–15 further compare area and power under identical frequency, showing that \textit{CorrectHDL} produces HDL designs with significantly lower area and power consumption than HLS-generated HDL designs.

\begin{figure}[t]
\vspace{-0.5cm}
\centering	
\includegraphics[width=1\linewidth]{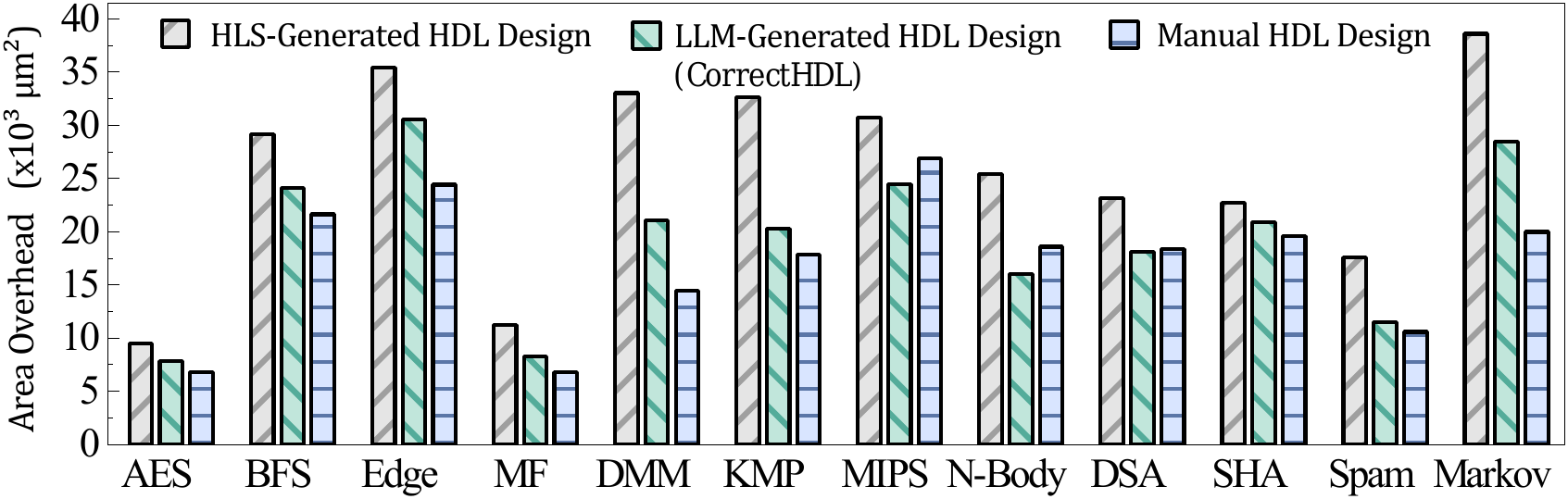}
\vspace{-0.8cm}
\caption{\fontsize{8.8pt}{9pt}\selectfont Comparison of area overhead.}
\label{fig:area}
\vspace{-0.2cm}
\end{figure}

\begin{figure}[t]
\vspace{-0cm}
\centering	
\includegraphics[width=1\linewidth]{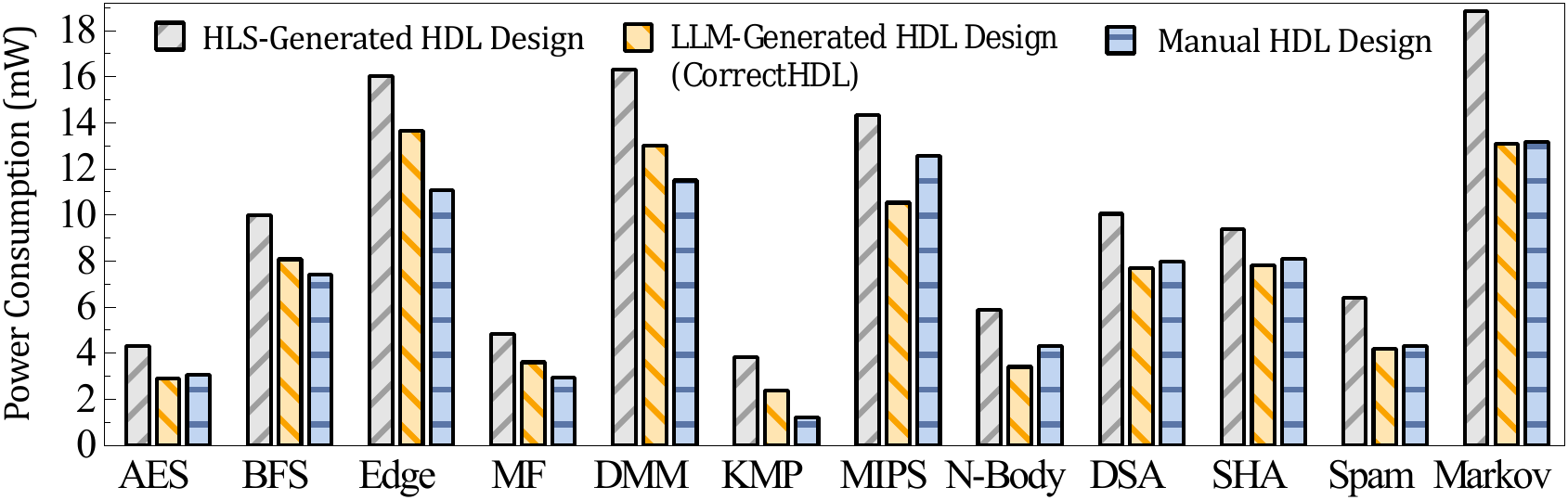}
\vspace{-0.8cm}
\caption{\fontsize{8.8pt}{9pt}\selectfont Comparison of power consumption.}
\label{fig:power}
\vspace{-0.5cm}
\end{figure}

To demonstrate the effectiveness of the rule-based C/C++ decomposition in \textit{CorrectHDL}, we compare the proposed decomposition method against direct LLM-based decomposition. As shown in Fig.~\ref{fig:cdecom}, the proposed approach achieves an average improvement of 18.19\% in decomposition pass rate. To assess the effectiveness of syntax repair for LLM-generated HDL designs, we compare the compilation pass rate achieved by \textit{CorrectHDL} (with RAG) against directly using the LLM. As shown in Fig.~\ref{fig:ragre}, \textit{CorrectHDL} improves the compilation pass rate by an average of 15.49\%.

To evaluate the quality of functional repair of submodules with \textit{CorrectHDL}, Fig.~\ref{fig:funcpr} compares the functional simulation pass rates of submodules using \textit{CorrectHDL}, using the LLM directly, and using the LLM with step-wise reasoning for functional repair. Step-wise reasoning improves the pass rate by an average of 5.90\% over direct LLM usage, while \textit{CorrectHDL}, using HLS as a functional reference, further improves the pass rate by an average of 28.05\%.

To demonstrate the top-level functional simulation pass rate of the HDL design with \textit{CorrectHDL}, which incorporates submodule feedback from the HLS-generated reference during the final HDL design generation, it is compared with three other settings, as shown in Fig.~\ref{fig:interpr}. \textit{Direct LLM Baseline:} The LLM is used directly to generate and debug the top-level HDL design. \textit{LLM with task decomposition:} The C/C++ program is decomposed into LLM-friendly submodules to generate and debug the top-level HDL design. \textit{\textit{CorrectHDL} with Only Top-Design Feedback:} Submodule-level feedback is disabled. Fig.~\ref{fig:interpr} shows that, by incorporating an HLS reference, task decomposition, and submodule boundary instrumentation, \textit{CorrectHDL} outperforms the three settings with average gains of 38.54\%, 30.73\%, and 9.35\% in functional simulation pass rate, respectively.

To demonstrate the area and power efficiency of HDL designs generated by \textit{CorrectHDL}, we compare them with those generated by the traditional HLS tool and by manual design from open-source implementations~\cite{b42, b43}.  For fairness, all HLS designs are compiled using \texttt{\#pragma design\_goal area} to explicitly target minimum area, while the open source designs are adapted so that their bit widths, array sizes, and other parameters are consistent with the other two approaches. As shown in Fig.~\ref{fig:area} and Fig.~\ref{fig:power}, area overhead and power consumption are evaluated under identical frequency. The synthesized circuits generated by \textit{CorrectHDL} achieve an average area reduction of 24.83\% and power reduction of 26.98\% compared with HLS-generated designs. Moreover, the HDL designs generated by \textit{CorrectHDL} approach the quality of human-engineered circuits in many cases, while for a few benchmarks, \textit{CorrectHDL} produces even better designs than those written by human engineers.

\vspace{0.14cm}
\section{Conclusion}\label{sec:fifth}
In this paper, we have proposed \textit{CorrectHDL}, an agentic LLM-assisted HDL design framework that leverages HLS-generated HDL as a functional reference. Starting from a C/C++ program, complex algorithms are decomposed into LLM-friendly submodules whose HDL implementations are generated by the LLM. Syntax errors are corrected via RAG, and functional discrepancies are resolved through differential verification against the HLS-generated golden reference.
Experimental results on 12 real-world benchmarks show that \textit{CorrectHDL} significantly improves both syntax and functional pass rates of LLM-generated HDL, while achieving lower area and power than HLS-generated designs and approaching the efficiency of manually crafted designs. Future work will integrate C/C++ program generation from natural language to establish a complete LLM-assisted workflow from natural language to HDL design. The LLM-generated circuits will also be optimized further to explore their potential in matching and even surpassing human-engineered designs.


\end{document}